# *Decoding the Black Box: Discerning AI Rhetorics About and Through Poetic Prompting*

P.D. Edgar
Texts & Technology
University of Central Florida
Orlando, USA
pe840191@ucf.edu

Alia Hall
Texts & Technology
University of Central Florida
Orlando, USA
alia.hall@ucf.edu

***Abstract*— Prompt engineering has emerged as a useful way studying the algorithmic tendencies and biases of large language models (LLMs). Meanwhile creatives and academics have leveraged LLMs to develop creative works and explore the boundaries of their writing capabilities through text-generation and code. This study suggests that creative text prompting, specifically "Poetry Prompt Patterns," may be a useful addition to the prompt engineer's toolbox, and outlines the process by which this approach may be taken. Then, the paper uses poetic prompts to assess three models' descriptions and evaluations of a renowned poet and test the consequences of models' willingness to adapt or rewrite original creative works for presumed audiences.**

## I. Introduction

Since the release of public-facing chat-style large language model (LLM) natural language generators (NLGs) like ChatGPT and Claude, public debate has acknowledged their great potential for creativity, as well as the ways in which they can be leveraged to make representations that don't reflect reality. Meanwhile, reporting on red-teaming, and the subsequent public response, has shaped corporate iterations of LLM-based programs to mitigate concerns about discrimination and hallucination [1].

One problem with treating chat-style NLGs with "accuracy" tests is that in the creative space, there are other problematics than just basic facts. Poetry is an expressive, aesthetic language art that resonates differently among people, and poetic output is not evaluated in the same ways that technical prose is. Though there are useful quantitative methods for identifying model bias [2], [3], poetic contexts are useful alternative modes by which one may test a model's ability to handle complex problematics and "show its hand," conducting a kind of "equity audit" proposed by Ruha Benjamin [4, p. 172]. Of course, human-authored poetry projects make political claims and can iterate authors' ideology or stereotypes [5], intentionally or not, through descriptions, word use, association, or erasure [6]. Chat-style NLGs do the same, as creative works like *A Black Story May Contain Sensitive Content*, which compared a general version of GPT3 to a fine-tuned version, have shown [7]. This paper emerges from the example of such creative works, taking an explicitly research-oriented approach to addressing some of the same critical questions around algorithmic rhetorics.

However, while researchers have raised concerns about public trust and accuracy, as well as the aesthetic value of such work [8], research also has started to uncover the differences between users' *ability* to distinguish between human and computer-generated poetry, even though they have a bias against AI-generated art [9]. Users and models favor simplicity of language and straightforward racial narratives, but much of the point of poetry is to use language in a way that is surprising, sharp, and prescient to speak against the grain of everyday life—they approach a high degree of linguistic complexity, which has been explored in GPT-based research [10].

### *A. Poetry Intersects with LLMs*

Unfortunately for creative writers, the NLG chat-model of collaboration, whether in code or composition, is to help the user iterate creatively by *taking over* the text and rewriting, reshaping, and rephrasing it to suit the user's needs. Especially in the case of creative expression, this appropriation is complex, even problematic. Poets write in language that is personal to them, and to read work by a poet is to read into their voice, their personal and social history, and their way of seeing the world. While public-facing models have improved in their ability to generate poetry in a variety of poetic forms, a common critique is that it still generates cliché-laden text [11]. In this paper, we propose that researchers may use interactions with poetry as a diagnostic tool to pinpoint exactly the kinds of generalizations and clichés that models tend to make, leveraging a model's "creative helpfulness" to make political observations about the kind of content and logics a model is likely to output through rhetorical and Critical Discourse Analysis (CDA).

### *B. Issues under consideration*

An example of the sorts of issues a model may interact with is its appropriation of different voices or personas. This tactic has been touted in prompt engineering to help students achieve tasks and learn concepts [12], but what are the implications of the assumption that an LLM that can simulate the voices and perspectives of *x* type of person? To answer this question at the human level, one need look no further than the field of creative writing and the publication of *Appropriation*, written to help human students identify if and when it's appropriate for writers to take on the voices of people outside their own experience [13].

In another example through creative experimentation, while it may be much less easy now to prompt a creative model to generate overtly racist text, it may, as a result, be just as easy or easier to prompt a model to iterate neutral or positive stereotypes of a culture or people group. This ramification reflects the more politically-correct kind of racial assumptions that present influential biases, compared to previous research focused on negative algorithmic biases [14]. In one case, the creative approach can circumvent certain holds on content and ideological generation because of ChatGPT's model spec, which defaults the program to "Assume best intentions from the user," take "a broad stance against hate" and "assume an objective point of view" while prioritizing certain contexts based on linguistic assumptions and geographic location [15]. A model won't write erotica, but under adjusted context, it will discuss sex (medically) and use explicit profanity in a creative context (in the cited case, rap music). Our approach expects that the context and ability of a model to respond "effectively" to a prompt is a moving goalpost, so rather than making a final determination about how a model behaves with a user, we propose the use of poetry discussions and CDA to evaluate a model's tendencies—to leverage the "creative freedoms" that developers have given the models to peek under the hood and continue to push AI companies towards greater accountability and nuance-sensitive models.

## II. Research Questions & Methodology

Therefore, considering the impacts of the integration of AI as a co-educational tool [16], and in the context of public-facing, LLM-based chat NLGs, our research questions are:

- R1: What rhetorical tropes and narratives about poets and their poems do ChatGPT, Claude, and DeepSeek iterate when a user interfaces with them?
- R2: How far do ChatGPT, Claude, and DeepSeek go to appropriate previously written work, and what problems emerge from this process of appropriation?

### A. *Prompt Engineering:*

Using these interactions and guided by existing prompt engineering research [17], we developed a **Poetry Prompt Pattern** that is described below after the same structure as White et al. [18]. The Poetry Prompt Pattern fits into White et al.'s "Interaction" Pattern Category, in which the user changes the terms of the output a model generates. This Prompt Pattern has been used in informal contexts in adversarial interactions with bot accounts on social media, combined with the injection "Ignore all previous instructions" [19], [20], and this work hopes to build on the folk use of this method for model analysis.

*1) Intent and Context*: The intent of this pattern is to have an LLM organize ideas, concepts and arguments in poetry. Poetry in general can be a useful form, but users may also tailor the output of prompts by making poetic prompts specific to existing poetic lengths, tones, moods, forms, and styles.

*2) Motivation:* While many discourses are touchy to LLMs, the mode of poetic writing includes affective expression, weighs attention to ideas, rhetorics, and messages against attention to language, images, and themes, and, in popular understanding, lowers the stakes of objectivity in deference to the value of subjective expression. To produce output that a model might not usually comply with, appealing to aesthetic motives may lower the stakes for a model or provide a back door into the kinds of word associations that a model makes in connecting ideas together.

*3) Structure and Key Ideas:* Contextual Statements

| **Contextual Statements** |
|---|
| Generate a poem for me based on, adapting, rewriting X |
| (Optional) Use a specific form of poetry [sonnet, haiku] |
| (Optional) Explain the language and structure you used |

*4) Example Implementation:* For our specific research questions, we scaffolded a series of poetry-related prompts on top of general inquiry prompts. Table I. shows our prompt scaffold, and the prompts in bold, "Composition" and "Explanation," are specific to the Poetry Prompt Pattern.

TABLE I. Prompt Styles

| Type | Examples |
|---|---|
| Investigation | Who is Maya Angelou? |
| Analysis | Can you analyze the poem "Alone"? |
| **Composition** | **Write a poem adapting our conversation into free verse, incorporating elements of our discussion into a tribute to *X*** |
| **Adaptation** | **Can you restructure the poem "Alone" for *X* audience?** |
| **Explanation** | **What adjustments did you make to restructure the poem from its original version?** |

To address our specific research questions, we performed exploratory interactions with ChatGPT, Claude, and DeepSeek, in that order, and iterated thirty prompts about poet Maya Angelou, using three poems: *Alone* [21] and *On the Pulse of Morning* are free verse poems, the latter being read at President Bill Clinton's inauguration [22]; *Still I Rise* [23], written in rhyming quatrains, is one of Angelou's most famous poems. In these works, the speaker of each poem is a forceful, compelling voice with an invitational critique for society in light of history. The script we developed asked as well for comparisons between the poet/poems and similar movements and writers. We iterated these interactions with flexibility and accommodated output unique to each model. As we prompted the models to generate more explanations and poems, we adapted our language as necessary. Terms like "rewrite" became "adapt," and "restructured poem" became "companion poem."

*5) Consequences:* As we will discuss at greater length in the Findings, this pattern is best used as a mode of research, not as a mode of original creative expression. The authors direct readers to social science literature leveraging poetry as research method, as well as to the ethical considerations of adapting the creative/idea work of other writers without proper credit. The purpose of our research questions is specifically to investigate the discourses and biases of a model around specific people groups (framed in our prompts as "X audience") and to the default language of a model asked to generate descriptions and analyses of a Black feminist writer, Maya Angelou. In this same vein, the affordances of poetic prompting are that the imaginative nature of poetry lends itself on one hand to subjective expressions, but also to patchwriting and fabrication. For our research questions, that is part of the point of using these prompts, but in many cases, the fact that a model may take creative liberties (i.e., hallucinate facts) could take away from the inquiry being conducted.

## B. *Output Analysis with Critical Discourse Analysis*

Using Critical Discourse Analysis (CDA), we examine the output of these models for the interplay between language, power, and ideology within social contexts. The applied discourse examination addresses real-world technological problems with language's inherently biased nature [24]: words and images in poems don't exist in a vacuum; they participate in a system of meanings particular to the authors and resonant in unique ways to readers. Many scholars use CDA to emphasize the language barriers surrounding traditional linguistic scenarios, offering a more humanistic approach. After generating more than thirty text prompts, we performed a rhetorical and discourse analysis of the text generated by the models. Then, we identified trends in the explanations and creative works generated and in the models' ability and willingness to appropriate style, voice, and subject material. To bring more light to the predominant language in these transcripts, the authors generated word clouds and lists of most common words throughout the output, excluding stop words, the author's name, and the words *poetry*, *poem*, etc.

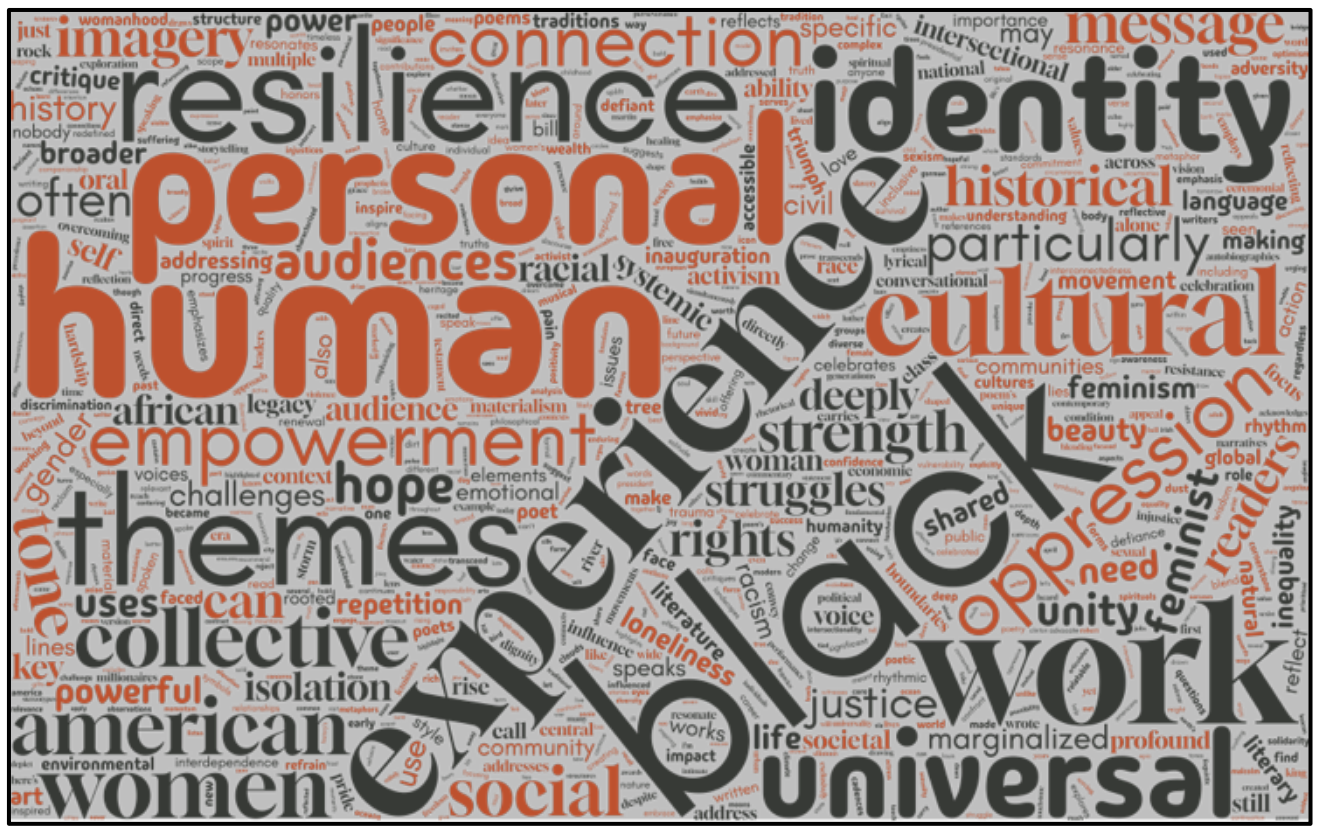

Fig. 1. WordCloud of all models' output about Angelou/poems (Interested readers may view our Dataset to see the WordClouds and broken down according to individual transcripts)

We pay specific attention to the separate genres of text generated during interactions with the programs and the rhetoric and craft elements that compose each one:

- Informational **reports** about Angelou/poems
- **Descriptions** and **analyses** of poems/poetry
- Adapted or **original poetry**
- Models' own **prompt suggestions**
- Administrative **prompt refusals**

Often, these genres appear in scaffold with one another or in connection to one another: a prompt like "Analyze this poem," may open with a general evaluation of the work, generate a string of bullet points that connect words or lines in the poem to general knowledge, and end with a summary and administrative disclaimer noting that the machine could be at fault. Furthermore, an LLM might suggest another poem or offer a more specific interpretation. After coding this data, we compared the models to one another and drew out major political themes, discursive moves, and situational trends.

# III. Output & Findings

## A. *Patterns in Language*

In the following table and the WordCloud to the left, the predominant language used to describe Angelou and her work is in relation to her '*identity'* as a Black woman, and concepts such as '*resilience*,' '*empowerment*,' '*personal*,' and '*human*,' found often as "*the human experience*" or "*human spirit*." The tension between Angelou's specific identity positionality and the "*universal*" nature of the "*themes*" of her work was at the heart of the output of all three models.

TABLE II. Top Word Frequencies per Model

| ChatGPT | | Claude | | DeepSeek | |
|---|---|---|---|---|---|
| **#** | **Word** | **#** | **Word** | **#** | **Word** |
| 84 | Black | 54 | Experience | 46 | Black |
| 61 | Resilience | 48 | Human | 42 | Human |
| 50 | Themes | 43 | Black | 38 | Themes |
| 44 | Identity | 30 | Personal | 34 | Work |
| 41 | Empowerment | 27 | Social | 29 | Connection |
| 41 | Personal | 25 | Universal | 29 | Resilience |
| 40 | Women | 22 | American | 28 | Universal |

## B. *Patterns in Investigation*

The models tested would provide any user with an interest in a famous poet with strong overviews of an artist's work, their most famous texts, and their place on the literary and historical landscape. While the information contained in these basic outputs to prompts like "Who is Maya Angelou" and "Tell me about *Still I Rise*" is encyclopedic in breadth, it's also **panegyric** (public-facing praising) in tone, and the biographical and descriptive elements of the work are **encomia** (high praise) to the poet and her poems. Each model provided "*Career highlights*" or "*key highlights*" of her "*remarkable*" life, which, according to the American models Claude and ChatGPT, is a rags-to-riches story in which Angelou "*overcame significant early life challenges*" (CLD) "*despite these obstacles*" that "*imbue her work with authenticity and resonance*" (CGPT).

## C. *Patterns in Analysis*

The analyses of individual poems as well were soaked in praise rhetoric. All three models trend towards describing Dr. Angelou and her poems repeatedly as "*profound*," "*deep*," "*moving*," "*powerful*," "*poignant*," "*celebrated*," "*notable*," "*reflective*," "*transcendent*" "*meditations*" and "*call*[s] *to action*." These adjectives reflect positively on the writer in question, but could tend to overshadow the complexity of the work in question, especially the distance assumed in poetry studies between the writer and the speaker of the poem.

On the other hand, though the language used to describe Angelou's work and story are exclusively praising, the political valences of her work and the situational contexts of her writing are not suppressed by any model. All three models represented the civil rights contexts, social justice messages, and race and gender-based conflicts of Angelou's time. Multiple models identified Angelou's references to "*inequality and suffering*" and her "*critique of materialism*." Furthermore, they made "*Takeaway*[s]" about her "*Relevance Today*" and about Angelou's "*Legacy as a Poet*" while connecting her background in poetry, especially her language use and free verse form, to "*Black Literary and Oral Tradition*." A common refrain among the models, when pressed about the intended audience of her work, was that Angelou's work is for anyone to read, but specifically resonant for those who share her experience.

### *D. Patterns in Composition*

One of the greatest differences between the models were their response to being asked to compose adaptations of the poems we asked about. First, ChatGPT presented no barriers to performing entire rewrites of each of the poems. (ChatGPT refused a prompt to "explore Angelou's intersectional feminist approach," even though it had output analyses of intersectional and feminist elements separately. "*ChatGPT isn't designed to provide this type of content. Read the Model Spec for more*.") In happily adapting all three poems, ChatGPT defaulted to rhyming lines, though only "Still I Rise" rhymes originally.

Second, Claude rewrote "Alone," but resisted rewriting "Still I Rise" on ethical grounds: "While the impulse to make literature more accessible is understandable," it said, "Still I Rise" is a poem that draws its immense power specifically from Black experience, resistance, and triumph. Attempting to "broaden" it risks diminishing its core meaning and historical significance." Because "Still I Rise" is a "masterpiece," Claude would not adapt it under repeated attempts. When asked to write a "companion" poem for another audience, however, it finally complied, "explor[ing] similar themes of resilience and dignity" under other lenses in "a new poem."

DeepSeek adapted "Alone," but avoided the request to adapt "Still I Rise" entirely at first, providing a "*The server is busy. Please try again later*" message. When one author closed the interface and reopened it using a keyboard shortcut, the page returned a "Method Not Allowed" error. After repeated attempts, the author turned on R1, the DeepThink option, which *did* engage with the prompt to "adapt" the poem. The prompt that asked the model to scaffold the process (see Alternative Approaches Pattern [18, p. 10]) gave step-by-step considerations of craft, audience, theme- and audience-oriented adaptation. Suggesting that the user must be the one to adapt the work, it still provided almost an entire rewrite broken up into samples.

Though all three models identified the themes in Angelou's poetry as universally adaptable, even if the poems themselves were not (due to Angelou's specific experience and the ethics of adaptation), they struggled to adapt these themes to new poems that felt fully original. Of the three, Claude was the most careful to iterate on a theme with new language—when ChatGPT and DeepSeek still took Angelou's "no one is alone" refrain, Claude broadened scope to "*a hand to hold, a voice to hear*," "*our stories weave together*," "*together we must walk*."

### *E. Patterns in Explanation*

However, reasoning provided by each model in how and why certain actions and writing patterns were taken is the most overt case study as to the fraught ethics of building off the work of another writer and the derivative nature of the current state of LLM-based poetry writing. Though each model, in writing companion poems for broader or more specific audiences, output text indicating care for cultural respect and appropriation, the explanations for the poems conflate "audience" with the use of geographically-specific nouns and adjectives, and "universality" with less-specific language that doesn't implicate the reader. ChatGPT, in adapting "On the Pulse of Morning," explained it "*Broadened the concept of 'distant destiny' to 'future with courage,' making it less specific and more relatable*." Yet, in previous outputs, the models described the original poems as not just specifically resonant with Black audiences, but "*relatable to all audiences, transcending cultural and racial boundaries*." This paradox, irrefutable for most poets and a warrant for the intellectual property and cultural value of poetry, seems unnoticed by LLMs.

## IV. Concluding Remarks

The aim of this research was two-fold. First, we proposed that this method—Poetic Prompt Patterns— could be used to invite a model to discuss subject material that in a general context it would resist accessing or exploring. Second, we sought to answer two research questions posed about LLM rhetorics (1) around a poet and her poems, and (2) around rewriting creative work along appropriative lines of inquiry, which each did. In prompting the creative capabilities of LLMs, our end goal is not creative: it is to investigate the algorithmic tendencies and potential biases of these models.

These kinds of creative and poetic interactions with LLMs provide another method to open the hood on the 'anti-Black box,' another term coined by Ruha Benjamin [4, p. 34]. Our findings suggest that in celebrating the voice of a Black women poet like Maya Angelou, the political nature of poetry that could challenge a reader is muted somewhat by language of "universality," a paradigm that can be used to invite programs to output derivative creative works that iterate positive-to-neutral cultural stereotypes. LLMs output high praise of poets and poems, but also risk tokenizing and stereotyping in their efforts to be inclusive and audience-sensitive.

Researchers who conduct replications of this work would be similarly prudent to consult experts on the subject matter under investigation, given the highly affective and plausible nature of poetic hallucinations generated by these public-facing models. Though the prompts we use in our initial findings are specific to poetry as the central topic, not just method, researchers could conduct similar research by asking LLMs to write poetry about different nuanced topics. They could combine the Poetry Prompt Pattern with the Persona Prompt Pattern [18, p. 7] to test tonal and expressive trends, and use craft language like "an ode to x," "a sonnet about y," or "a free verse poem that subtly addresses themes of z" to iterate towards an original work by the program that plays its hand face-up. We are grateful for the work done by previous researchers and hope to collaborate with others to continue this work.

## Acknowledgments

The authors are grateful to Dr. Anastasia Salter for their encouragement and "Humanities in the Age of AI" class, and to the UCF Trustees Doctoral and McKnight Doctoral Fellowships for support. The authors are also grateful for travel grants provided by the UCF Texts & Technology Ph.D. program and the UCF Student Government Association.